\documentclass[journal]{IEEEtran}
\usepackage{amsmath}
\usepackage{algorithm}
\usepackage{algorithmic}
\usepackage{graphicx}
\usepackage{multirow}
\usepackage{tabularx}
\usepackage{xcolor}
\usepackage{array}
\usepackage{amssymb}
\usepackage{bbding}
\usepackage{subfigure}
\usepackage{diagbox}
\usepackage{rotating}
\usepackage{bm}
\usepackage{tikz} 
\usepackage{booktabs}
\usepackage{cite}
\usepackage{CJK}
\usepackage{overpic}
\usepackage[colorlinks,
            linkcolor=red,
            anchorcolor=blue,
            citecolor=green
            ]{hyperref}

\newcommand{\addFig}[1]{}
\newcommand{\addFigs}[1]{}

\usepackage{subfigure}
\newcommand{\etal}{\textit{et~al}.~}
\newcommand{\ie}{\textit{i}.\textit{e}.,~}

\usepackage{balance}
\usepackage{cleveref}
\crefformat{section}{\S#2#1#3} 
\crefformat{subsection}{\S#2#1#3}
\crefformat{subsubsection}{\S#2#1#3}

\ifCLASSINFOpdf
\else
\fi

\begin{document}
%
\title{RGB-T Semantic Segmentation with Location, Activation, and Sharpening}
%
%
%

\author{Gongyang~Li,
	Yike~Wang,
	Zhi~Liu,~\IEEEmembership{Senior Member,~IEEE},\\
	Xinpeng~Zhang,~\IEEEmembership{Member,~IEEE},
	and~Dan~Zeng,~\IEEEmembership{Senior Member,~IEEE}

\thanks{All authors are with Key Laboratory of Specialty Fiber Optics and Optical Access Networks, Joint International Research Laboratory of Specialty Fiber Optics and Advanced Communication, Shanghai Institute for Advanced Communication and Data Science, Shanghai University, Shanghai 200444, China, and School of Communication and Information Engineering, Shanghai University, Shanghai 200444, China (email: ligongyang@shu.edu.cn; shuwangyike@shu.edu.cn; liuzhisjtu@163.com; xzhang@shu.edu.cn; dzeng@shu.edu.cn).} 
\thanks{\textit{Corresponding authors: Zhi Liu and Xinpeng Zhang.}}
}

\markboth{IEEE TRANSACTIONS ON CIRCUITS AND SYSTEMS FOR VIDEO TECHNOLOGY}%
{Shell \MakeLowercase{\textit{et al.}}: Bare Demo of IEEEtran.cls for IEEE Journals}

\maketitle

\begin{abstract}
Semantic segmentation is important for scene understanding.
To address the scenes of adverse illumination conditions of natural images, thermal infrared (TIR) images are introduced.
Most existing RGB-T semantic segmentation methods follow three cross-modal fusion paradigms, \ie encoder fusion, decoder fusion, and feature fusion.
Some methods, unfortunately, ignore the properties of RGB and TIR features or the properties of features at different levels.
In this paper, we propose a novel feature fusion-based network for RGB-T semantic segmentation, named \emph{LASNet}, which follows three steps of location, activation, and sharpening.
The highlight of LASNet is that we fully consider the characteristics of cross-modal features at different levels, and accordingly propose three specific modules for better segmentation.
Concretely, we propose a Collaborative Location Module (CLM) for high-level semantic features, aiming to locate all potential objects.
We propose a Complementary Activation Module for middle-level features, aiming to activate exact regions of different objects.
We propose an Edge Sharpening Module (ESM) for low-level texture features, aiming to sharpen the edges of objects.
Furthermore, in the training phase, we attach a location supervision and an edge supervision after CLM and ESM, respectively, and impose two semantic supervisions in the decoder part to facilitate network convergence.
Experimental results on two public datasets demonstrate that the superiority of our LASNet over relevant state-of-the-art methods.
The code and results of our method are available at https://github.com/MathLee/LASNet.
\end{abstract}

\begin{IEEEkeywords}
RGB-T semantic segmentation, discriminative treatment, collaborative location, complementary activation, edge sharpening.
\end{IEEEkeywords}

\IEEEpeerreviewmaketitle

\section{Introduction}
\IEEEPARstart{S}{emantic} segmentation~\cite{2015FCN,2015DeconvNet,DeepLabV1,17SegNet}, also known as scene parsing, is a fundamental topic in computer vision, and focuses on assigning a category label to each pixel in a natural image, which is a dense prediction task.
It plays an important role in autonomous driving~\cite{1autonomous,2autonomous}, medical analysis~\cite{2015Unet}, remote sensing scene understanding~\cite{1remote} and so on.
In the past decade, many effective solutions~\cite{2017PSP,2019CCNet,2021Segmenter,2019DANet,2020WSSS,2021Segformer,2022SAFA,2021CIPC,2022rethinking} for semantic segmentation have been proposed.
However, there are still some hard scenes that unimodal image-based methods cannot handle well, such as cluttered backgrounds, adverse illumination conditions (even darkness), and occlusions by fog or smoke.
Therefore, researchers additionally introduce thermal infrared (TIR) images, and combine them with natural images to handle the above hard scenes, forming an emerging task called RGB-T semantic segmentation~\cite{2017MFNet,2020PSTNet}.
TIR images reflect the surface temperature of objects collected by the TIR sensor, and are insensitive to illumination changes, making up for the shortcomings of unimodal natural images~\cite{VT821,2022ECFFNet,2022APNet,2022CGMDRNet,RGBTTracking}.

\begin{figure}[t!]
  \centering
  \footnotesize
  \begin{overpic}[width=1\columnwidth]{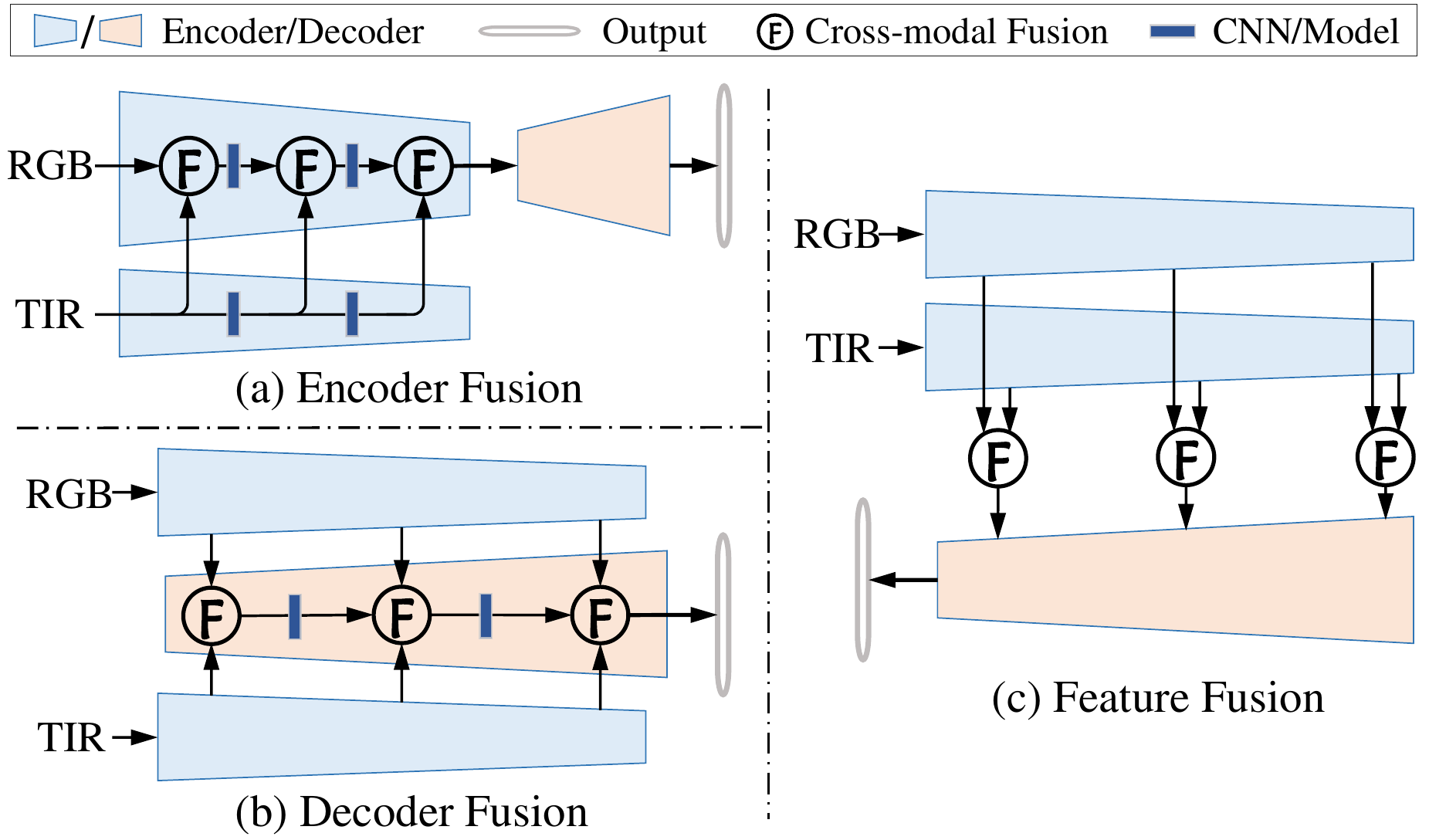}
  \end{overpic}
  \caption{Three typical cross-modal fusion paradigms for RGB-T semantic segmentation, including encoder fusion~\cite{2019RTFNet,2021MLFNet,2021FEANet,2021FuseSeg,2022MFFENet}, decoder fusion~\cite{2017MFNet,2021ABMDRNet,2022MTANet}, and feature fusion~\cite{2021GMNet,2022GCNet,2022EGFNet}.
    }\label{fig:example}
\end{figure}

%
For multi-modal inputs, the key is how to effectively make them complement each other.
Therefore, researchers strive to develop effective cross-modal fusion strategies for RGB images and TIR images in RGB-T semantic segmentation.
In general, as shown in Fig.~\ref{fig:example}, there are currently three typical cross-modal fusion paradigms for RGB-T semantic segmentation, including encoder fusion~\cite{2019RTFNet,2021MLFNet,2021FEANet,2021FuseSeg,2022MFFENet}, decoder fusion~\cite{2017MFNet,2021ABMDRNet,2022MTANet}, and feature fusion~\cite{2021GMNet,2022GCNet,2022EGFNet}.
Specifically, encoder fusion paradigm directly adopts element-wise summation to achieve cross-modal feature fusion on the feature extraction stage.
Decoder fusion paradigm usually adopts the same module to process cross-modal features at different levels on the inference stage.
Feature fusion paradigm focuses on fusing multi-level cross-modal features between encoder and decoder, and usually employs the same module.

The above three fusion paradigms have greatly promoted the development of RGB-T semantic segmentation.
However, the disadvantages of them are obvious.
The cross-modal feature fusion operation of the encoder fusion paradigm is the simple element-wise summation, ignoring the properties of RGB and TIR features.
The other two paradigms design specific modules to fuse cross-modal features, and have the same problem of using the unified module to process features at different levels.
In fact, features at different levels of convolutional neural networks (CNNs)~\cite{1989CNN} have unique characteristics.
The indiscriminate treatment ignores the properties of features at different levels and is suboptimal.

Driven by the aforementioned observation, in this paper, we propose a novel \emph{LASNet} for RGB-T semantic segmentation, aiming to explore the characteristics of features at different levels and to mine the common valuable content contained in cross-modal features.
We implement LASNet through the feature fusion paradigm.
Different from previous methods, we achieve accurate segmentation via three steps of object location, region activation, and edge sharpening.
We arrange these three steps reasonably according to the characteristics of features at different levels.
Furthermore, we design three dedicated modules for these three steps instead of the unified modules used by previous methods.

Specifically, we divide the basic cross-modal features extracted by the backbone into three levels, \ie high, middle, and low levels.
For high-level features, we propose a Collaborative Location Module (CLM), which builds pixel-level correlations~\cite{2022COSNet,2022CorrNet} among semantic representations to locate all potential objects.
For middle-level features, we propose a Complementary Activation Module (CAM), which builds on the attention mechanism~\cite{2018CBAM,2019DANet} to generate informative features for exact region activation.
For low-level features, we propose an Edge Sharpening Module (ESM), which extracts edges using multiple receptive fields to sharpen objects of different sizes.
In addition, we impose the location supervision on CLM and the edge supervision on ESM to improve the accuracy of object location and edge extraction.
In this way, the above three modules make good use of different levels of features, facilitating the proposed LASNet to generate satisfactory segmentation results.

Our main contributions are summarized as follows:
\begin{itemize}
\item We fully explore the characteristics of cross-modal features at different levels, and propose a novel \emph{LASNet} for RGB-T semantic segmentation, following three steps of location, activation, and sharpening.

\item We propose specific CLM, CAM, and ESM for high-, middle-, and low-level cross-modal features, respectively, which are responsible for object location, region activation, and edge sharpening.
These three modules mine the common valuable content contained in cross-modal features at different levels to generate discriminative features, facilitating accurate segmentation.

\item We evaluate the proposed LASNet on MFNet dataset and PSTNet dataset.
The results demonstrate that our LASNet outperforms state-of-the-art methods and show the reasonableness of discriminative treatment of cross-modal features at different levels in our LASNet.

\end{itemize}

The rest of this paper is organized as follows.
In Sec.~\ref{sec:related}, we review the related work of RGB and multi-modal semantic segmentation.
In Sec.~\ref{sec:OurMethod}, we elaborate the proposed LASNet.
In Sec.~\ref{sec:exp}, we present comprehensive experiments and analyses.
Finally, in Sec.~\ref{sec:con}, we conclude this work.

\section{Related Work}
\label{sec:related}

\subsection{RGB Semantic Segmentation}
\label{sec:Tra_ORSI_SOD}
Recently, with the help of CNNs, RGB semantic segmentation (\ie unimodal semantic segmentation) has made amazing progress.
Long~\etal\cite{2015FCN} proposed the milestone Fully Convolutional Network (FCN), which is the first end-to-end CNN-based semantic segmentation method.
Subsequently, Noh~\etal\cite{2015DeconvNet} and Badrinarayanan~\etal\cite{17SegNet} proposed the Deconvolution Network (DeconvNet) and the encoder-decoder architecture-based SegNet, respectively, which are two other pioneering works.
Inspired by the above works, many technologies have been developed and applied for RGB semantic segmentation, such as multi-scale strategy, contextual dependency, and transformers.

As a representative work of multi-scale strategy, Chen \etal proposed a series of methods, including DeepLabV1~\cite{DeepLabV1}, DeepLabV2~\cite{DeepLabV2}, DeepLabV3~\cite{DeepLabV3}, and DeepLabV3+~\cite{DeepLabV3+}.
They introduced atrous convolutions to capture multi-scale information.
Furthermore, they proposed atrous spatial pyramid pooling (ASPP), and made the network deeper with atrous convolutions.
Yang~\etal\cite{2018DenseASPP} proposed the Densely connected ASPP (DenseASPP) to generate features with very large receptive fields.
Zhao~\etal\cite{2017PSP} adopted four pooling layers of different sizes to capture the multi-scale global information, and proposed the Pyramid Scene Parsing Network (PSPNet).
Ji~\etal\cite{2021EDCC} adopted the spatial pyramid pooling to ensemble multi-scale features in the encoding stage, and learned boundary information in the decoding stage.

Some researchers focus on exploring contextual dependency for RGB semantic segmentation.
For example, Yuan~\etal\cite{2018OCNet} introduced the self-attention mechanism to learn the pixel-wise similarity maps to identify objects belonging to the same category as the target pixel.
Moreover, Yuan~\etal\cite{202OCP} built the pixel-region relation, and enhanced the representation of each pixel by the object-contextual representation for accurate segmentation.
Fu~\etal\cite{2019DANet} proposed Dual Attention Network (DANet) to model the global dependencies in channel and spatial dimensions via self-attention.
Huang~\etal\cite{2019CCNet} proposed the criss-cross attention to capture contextual information in an efficient way.
Similar to~\cite{2019CCNet}, Liu~\etal\cite{2020EfficientFCN} and Weng~\etal\cite{2022SAFA} focused on the efficiency of algorithms, and developed real-time and accurate segmentation methods.
To better mine global contextual dependencies, Strudel~\etal\cite{2021Segmenter} built the Segmenter on ViT~\cite{2021ViT}. 
Xie~\etal\cite{2021Segformer} proposed a cutting-edge Transformer framework, named SegFormer, with a lightweight All-MLP decoder.

Overall, semantic segmentation in unimodal images has achieved great success, but the above segmentation methods often fail in challenging scenes, such as low contrast, cluttered backgrounds, adverse illumination conditions, and occlusions by fog or smoke.
The researchers therefore introduce additional information to handle these difficult scenes, proposing multi-modal semantic segmentation.

\subsection{Multi-modal Semantic Segmentation}
\label{sec:CNN_ORSI_SOD}
In this paper, multi-modality data generally refers to RGB images and depth maps or TIR images, \ie RGB-D or RGB-T.
In addition to semantic segmentation, these two types of multi-modal data have a wide range of applications in computer vision, such as salient object detection~\cite{VT821,20CMWNet,20ICNet,21HAINet}, tracking~\cite{RGBTTracking}, and glass segmentation~\cite{RGBTGlass,RGBDGlass}.
Obviously, depth maps and TIR images are different and have their own characteristics.
Depth maps contain rich geometric distance information, while TIR images reflect the surface temperature of objects.

\begin{figure*}
	\centering
	\begin{overpic}[width=0.85\textwidth]{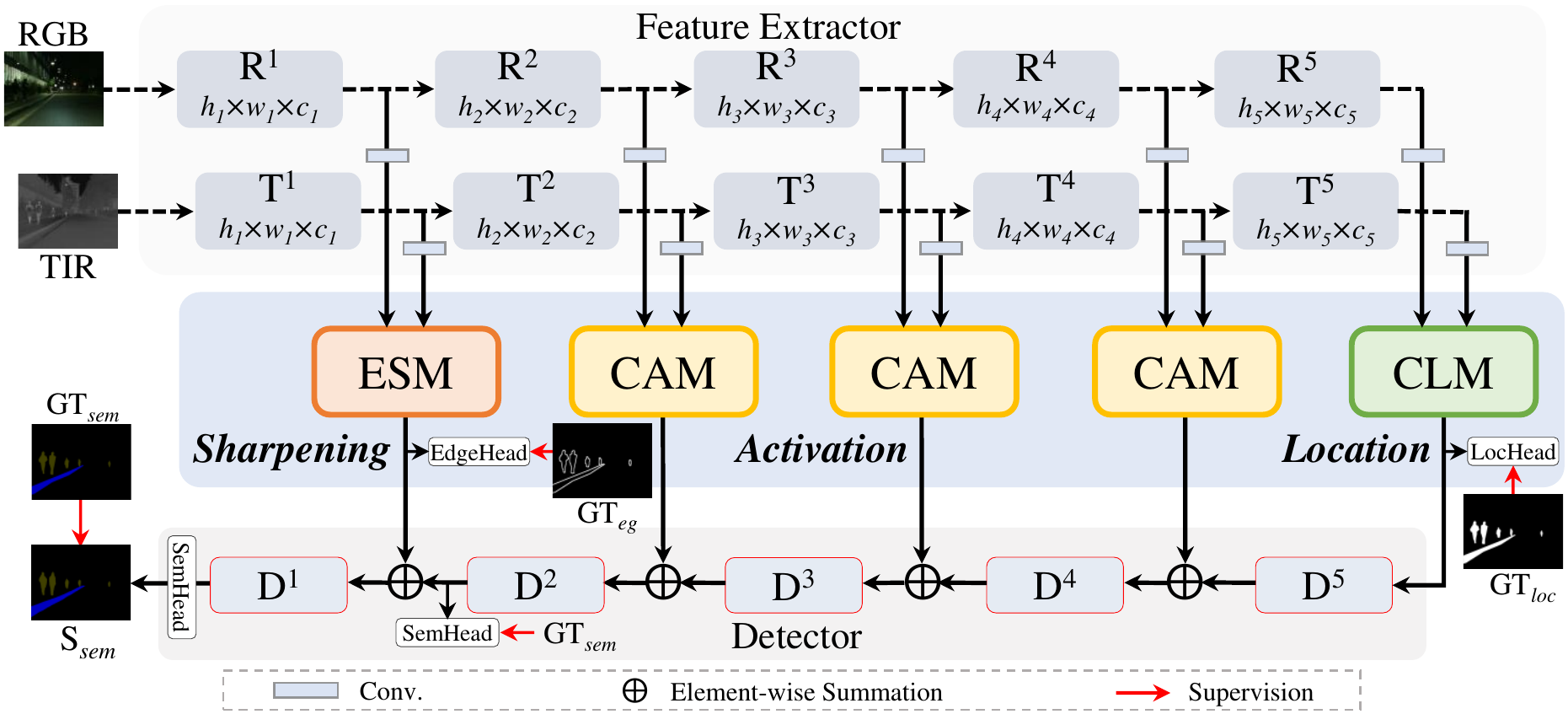}
    \end{overpic}
	\caption{Pipeline of the proposed LASNet.
	 Overall, LASNet follows the feature fusion paradigm, and consists of three parts, including a feature extractor, three specific modules, and a decoder.
	 Here, we adopt the ResNet-152~\cite{2016ResNet} as the backbone, and construct a parallel ResNet-152 structure for feature extraction, generating five-level basic cross-modal features.
	 Then, we arrange Collaborative Location Module (CLM),  Complementary Activation Module (CAM), and Edge Sharpening Module (ESM) for the high-, middle-, and low-level features, respectively.
	 CLM enhances the location representation of all potential objects.
	 CAM activates exact regions of different objects in three intermediate levels of features.
	 ESM sharpens the edges of objects.
	 Finally, according to the output features of the above three modules, we generate the segmentation result $\mathbf{S}_{sem}$ in the decoder.
    }
    \label{fig:Framework}
\end{figure*}

\textit{1) RGB-D Semantic Segmentation.}
Depth maps are useful for handling scenes with low contrast and disturbing backgrounds.
Cao~\etal\cite{2021ShapeConv} directly concatenated RGB image and depth map as the network input, and proposed the shape-aware convolutional layer to construct a segmentation network.
Hazirbas~\etal\cite{2016FuseNet} fused RGB features and depth features through element-wise summation in the two-stream encoder.
Differently, some researchers focus on the fusion of different levels of features.
For example, Wang~\etal\cite{2016LCSF} extracted RGB features and depth features independently, and then learned common features from the extracted high-level features to improve the segmentation accuracy.
Hu~\etal\cite{2019ACNet} proposed an attention complementary module to enhance the independently extracted features, and then adopted element-wise summation for simple fusion.
Moreover, Chen~\etal\cite{2020SA-Gate} proposed a separation-and-aggregation gate to fuse features at multiple levels, and integrated the fused features into the encoder.
There are other approaches that do not follow the above structures.
Wang~\etal\cite{2018D-CNN} proposed depth-aware convolution and depth-aware average pooling in a depth-aware CNN, integrating geometry into CNN.
Lin~\etal\cite{2020ZigZag} proposed the zig-zag architecture to construct context features of RGB image, and then matched the image region and discrete depth map to achieve the eventual segmentation.

\textit{2) RGB-T Semantic Segmentation.}
TIR images are useful for handling scenes with adverse illumination conditions and occlusions by fog or smoke.
We review existing RGB-T semantic segmentation methods according to the three paradigms in Fig.~\ref{fig:example}.
Sun~\etal\cite{2019RTFNet,2021FuseSeg} proposed the encoder fusion paradigm, which integrates TIR features into RGB features through element-wise summation and is similar to~\cite{2016FuseNet}.
Following this paradigm, Guo~\etal\cite{2021MLFNet} proposed a specific decoder with multi-level skip connections and the auxiliary decoding module for segmentation.
Deng~\etal\cite{2021FEANet} introduced a feature-enhanced attention module to fuse RGB and thermal information at multiple levels in the encoder part.
In the decoder, Zhou~\etal\cite{2022MFFENet} concatenated all features extracted from encoder, and proposed multi-label supervision to assist semantic segmentation.
Differently, Ha~\etal\cite{2017MFNet} proposed the decoder fusion paradigm, which adopts two independent networks to extract RGB features and TIR features and integrates them in the decoder.
Zhang~\etal\cite{2021ABMDRNet} reduced multi-modality difference in the independent encoders, and achieved cross-modal feature fusion via the channel weighted fusion module in the decoder.
Zhou~\etal\cite{2022MTANet} extracted semantic information from high-level RGB and TIR features via concatenation-attention operation, and proposed a hierarchical multimodal fusion module for multiscale fusion in the decoder.
Zhou~\etal\cite{2021GMNet,2022GCNet,2022EGFNet} separated the cross-modal feature fusion from the encoder and decoder, proposing the feature fusion paradigm.
They adopted the multi-modal fusion module to process RGB and TIR features at different levels, and extracted global information from the high-level fused features to guide the segmentation process of objects.

The above methods almost indiscriminately process features at different levels.
Only a few methods~\cite{2021GMNet,2022MTANet,2022GCNet,2022EGFNet} designed specific modules for high-level features, but these modules are based on some simple fusion strategies.
In this paper, we fully explore the characteristics of cross-modal features at different levels, and propose CLM, CAM, and ESM for high-, middle-, and low-level cross-modal features, respectively.
Our three modules are specially tailored to the characteristics of different levels of features.
Based on three specific modules and the feature fusion paradigm, we propose a complete solution, \ie LASNet, for RGB-T semantic segmentation.

\section{Proposed Method}
\label{sec:OurMethod}
In this section, we present the proposed LASNet in detail.
In Sec.~\ref{sec:Overview}, we introduce the network overview of LASNet.
In Sec.~\ref{sec:CLM}, Sec.~\ref{sec:CAM}, and Sec.~\ref{sec:ESM}, we elaborate the proposed CLM, CAM, and ESM, respectively.
In Sec.~\ref{sec:Loss Function}, we present the loss function.


\subsection{Network Overview}
\label{sec:Overview}
As depicted in Fig.~\ref{fig:Framework}, the proposed LASNet is based on the feature fusion paradigm, including a feature extractor, three specific modules, and a decoder.
For feature extractor, we adopt two parallel ResNet-152~\cite{2016ResNet} backbones, named RGB branch and TIR branch, to extract cross-modal features from RGB image and TIR image.
We denote the five convolution blocks in RGB branch and TIR branch as R$^{i}$ and T$^{i}$ ($i=1,2,3,4,5$), respectively, and the corresponding output RGB and TIR features as $\{\boldsymbol{\hat{f}}^{i}_{\rm r}, \boldsymbol{\hat{f}}^{i}_{\rm t}\} \in \mathbb{R}^{h_i\!\times\!w_i\!\times\!{c}_i}$.
The input size is denoted as $H\!\times\!W$, so $h_i$ is $\frac{H}{2^{i}}$, $w_i$ is $\frac{W}{2^{i}}$, and $c_i\in\{64,256,512,1024,2048\}$.
Here, the RGB branch and the TIR branch share parameters, which not only keeps the extracted features in the same feature space, but also reduces the number of parameters.
Furthermore, we employ two identical convolutional layers (\ie shared parameters) to project the same level of RGB and TIR features, $\boldsymbol{\hat{f}}^{i}_{\rm r}$ and $\boldsymbol{\hat{f}}^{i}_{\rm t}$, to $\boldsymbol{f}^{i}_{\rm r}$ and $\boldsymbol{f}^{i}_{\rm t}$, respectively, with fewer channels (\ie $c_i\in\{64,128,256,256,512\}$) to reduce computational cost.

We divide the extracted basic cross-modal features into three levels, that is, $\{\boldsymbol{f}^{5}_{\rm r}, \boldsymbol{f}^{5}_{\rm t}\}$ are the high level, $\{\boldsymbol{f}^{2\sim4}_{\rm r}, \boldsymbol{f}^{2\sim4}_{\rm t}\}$ are the middle level, and $\{\boldsymbol{f}^{1}_{\rm r}, \boldsymbol{f}^{1}_{\rm t}\}$ are the low level.
We arrange CLM, CAM, and ESM for these three levels of features to achieve object location, region activation, and edge sharpening, respectively.
Concretely, CLM can enhance the location representation of all potential objects with the help of location supervision.
CAM can activate multi-level exact regions of different objects.
ESM can extract edge information of objects with the help of edge supervision.
These three specific modules are the core components of our LASNet, implementing three important segmentation-friendly functions.
With the informative features generated by the above modules from the five-level features, we achieve the accurate segmentation result $\mathbf{S}_{sem}$ in a general decoder, which is composed of five decoder blocks, and each decoder block includes a dropout layer, two convolutional layers, and an upsampling operation.

\begin{figure}
\centering
\footnotesize
  \begin{overpic}[width=1\columnwidth]{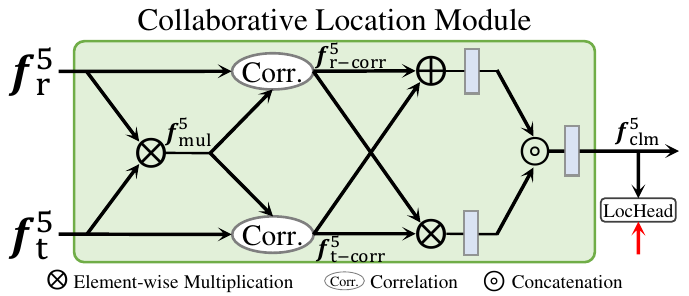}
  \end{overpic}
\caption{
Illustration of the Collaborative Location Module (CLM).
}
\label{CLM_structure}
\end{figure}

\subsection{Collaborative Location Module}
\label{sec:CLM} 
As we all know, high-level features contain rich semantic information, and have strong representations for object location.
And cross-modal RGB and TIR features are complementary, which is more conducive to locating objects.
Inspired by the object segmentation works which explore the correlation (\ie co-attention~\cite{2016CoAtt}) of target objects in consecutive video frames~\cite{2022COSNet} (\ie cross-frame) and features of successive levels~\cite{2022CorrNet} (\ie cross-level), we propose \emph{Collaborative Location Module} to model the pixel-level correlation of cross-modal high-level semantic features to collaboratively determine the object location.
In addition, we mine the valuable representation from two types of feature combinations (\ie summation and multiplication) in the CLM.
We illustrate the CLM in Fig.~\ref{CLM_structure}.
The input features of the CLM are $\boldsymbol{f}^{5}_{\rm r}$ and  $\boldsymbol{f}^{5}_{\rm t}$.
The whole process of CLM can be divided into cross-modal correlation modeling and correlation combinations.

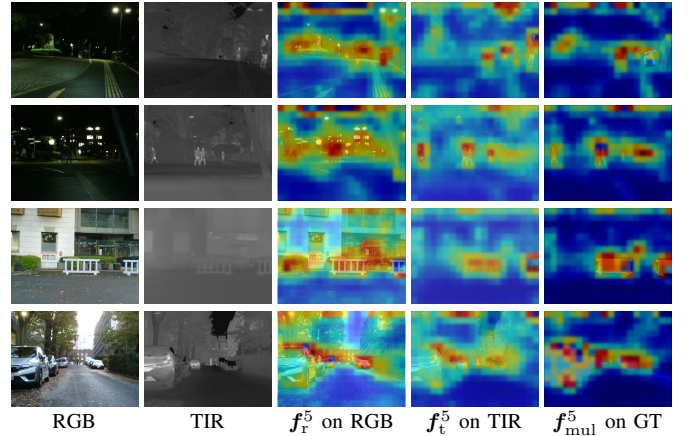
\begin{figure}
\centering
\footnotesize
  \begin{overpic}[width=1\columnwidth]{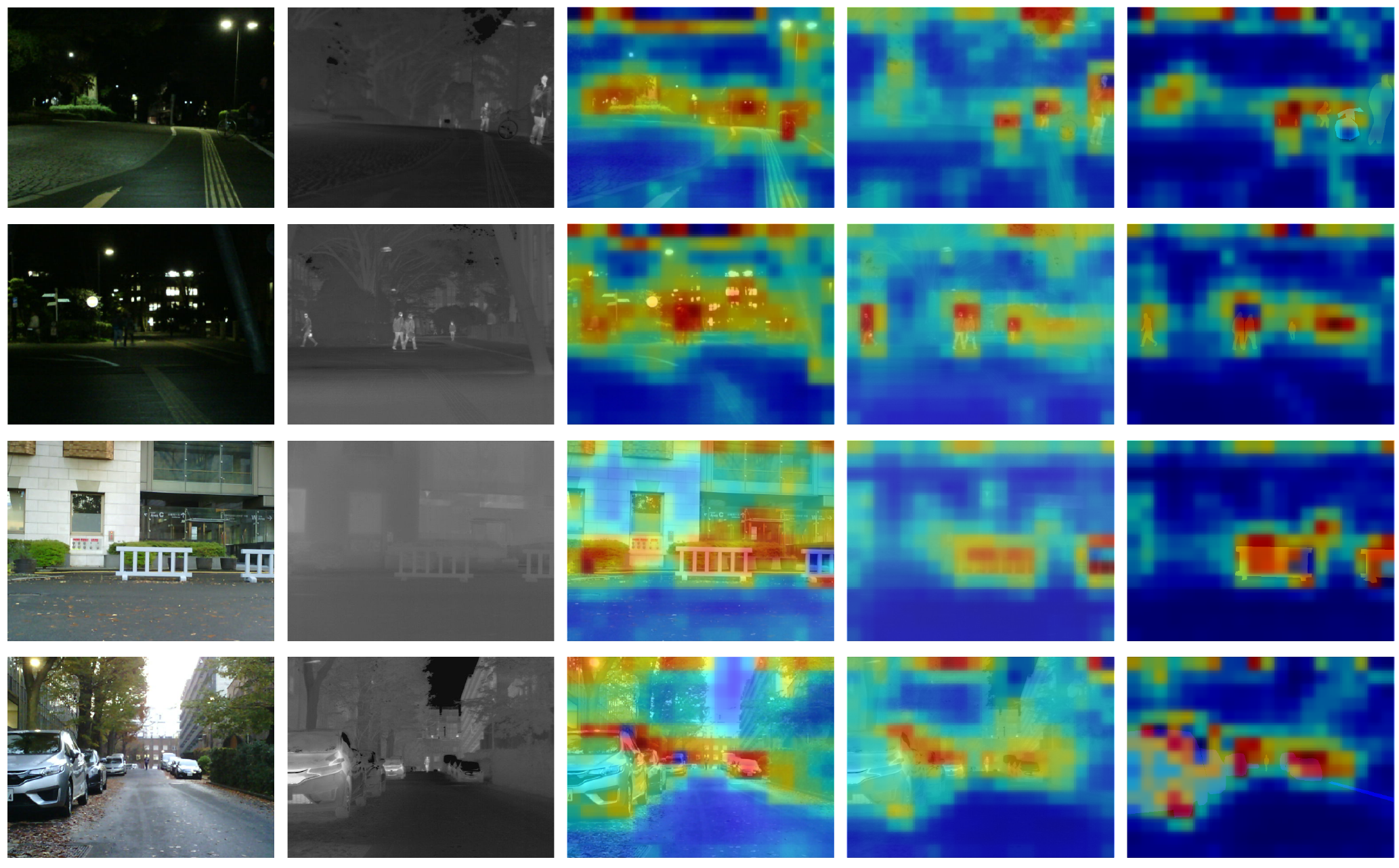}
    \put(5.6,-2.5){ RGB }
    \put(26.3,-2.5){ TIR}
    \put(41.0,-2.5){ $\boldsymbol{f}^{5}_{\rm r}$ on RGB }
    \put(61.6,-2.5){ $\boldsymbol{f}^{5}_{\rm t}$ on TIR }
    \put(80.58,-2.5){ $\boldsymbol{f}^{5}_{\rm mul}$ on GT }

  \end{overpic}
\caption{Feature visualization of $\boldsymbol{f}^{5}_{\rm r}$, $\boldsymbol{f}^{5}_{\rm t}$, and $\boldsymbol{f}^{5}_{\rm mul}$ in CLM. Here, we superimpose $\boldsymbol{f}^{5}_{\rm r}$, $\boldsymbol{f}^{5}_{\rm t}$, and $\boldsymbol{f}^{5}_{\rm mul}$ on RGB image, TIR image, and ground truth (GT), respectively. Please zoom-in for details.
}
\label{FeatureVisual}
\end{figure}

\textit{1) Cross-modal Correlation Modeling}.
In low illumination environments, objects in the RGB image are drowned in the background, as the input RGB image in Fig.~\ref{fig:Framework}, but are evident in the TIR image.
We believe that the multiplication operation can highlight drowned regions of RGB features with the evident content of TIR features in the product features, and simultaneously extract the coexistence information of two features.
Therefore, we first perform the element-wise multiplication operation on $\boldsymbol{f}^{5}_{\rm r}$ and  $\boldsymbol{f}^{5}_{\rm t}$, generating $\boldsymbol{f}^{5}_{\rm mul} \in \mathbb{R}^{h_5\!\times\!w_5\!\times\!{c}_5}$.
In Fig.~\ref{FeatureVisual}, we visualize $\boldsymbol{f}^{5}_{\rm r}$, $\boldsymbol{f}^{5}_{\rm t}$, and $\boldsymbol{f}^{5}_{\rm mul}$.
In the first two cases of nighttime, the persons and objects are almost invisible in RGB images, but are obvious in TIR images.
This makes the object regions in $\boldsymbol{f}^{5}_{\rm r}$ indistinguishable, but the object regions in $\boldsymbol{f}^{5}_{\rm t}$ can be easily highlighted, resulting in the object regions in the product features $\boldsymbol{f}^{5}_{\rm mul}$ are highlighted more accurately than those in $\boldsymbol{f}^{5}_{\rm r}$.
In the last two cases, the object regions in $\boldsymbol{f}^{5}_{\rm t}$ and $\boldsymbol{f}^{5}_{\rm r}$ are relatively clear.
Through the multiplication operation, the common objects in $\boldsymbol{f}^{5}_{\rm t}$ and $\boldsymbol{f}^{5}_{\rm r}$ can be extracted, resulting in that $\boldsymbol{f}^{5}_{\rm mul}$ can highlight the objects.

Then, we model the pixel-level correlation of $\{\boldsymbol{f}^{5}_{\rm mul}, \boldsymbol{f}^{5}_{\rm r}\}$ and $\{\boldsymbol{f}^{5}_{\rm mul}, \boldsymbol{f}^{5}_{\rm t}\}$, respectively, to collaboratively identify objects in cross-modal features, which can be formulated as follows:
\begin{equation}
   \begin{aligned}
    \boldsymbol{f}^{5}_{\rm r-corr} =   Corr(\boldsymbol{f}^{5}_{\rm mul}, \boldsymbol{f}^{5}_{\rm r}),  \\
    \boldsymbol{f}^{5}_{\rm t-corr}  =   Corr(\boldsymbol{f}^{5}_{\rm mul}, \boldsymbol{f}^{5}_{\rm t}),
    \label{eq:1}
    \end{aligned}
\end{equation}
where $\{\boldsymbol{f}^{5}_{\rm r-corr}, \boldsymbol{f}^{5}_{\rm t-corr} \} \in \mathbb{R}^{h_5\!\times\!w_5\!\times\!{c}_5}$ are the cross-modal correlation features, and $Corr(\cdot)$ is the co-attention operation~\cite{2016CoAtt}.
Here, taking $\boldsymbol{f}^{5}_{\rm mul}$ and $\boldsymbol{f}^{5}_{\rm r}$ as an example, the co-attention operation  computes the spatial dependencies of $\boldsymbol{f}^{5}_{\rm mul}$ and $\boldsymbol{f}^{5}_{\rm r}$ through the matrix multiplication, producing an affinity matrix, and then transfers the valuable information of the affinity matrix to $\boldsymbol{f}^{5}_{\rm r}$.

\textit{2) Correlation Combinations}.
$\boldsymbol{f}^{5}_{\rm r-corr}$ and $\boldsymbol{f}^{5}_{\rm t-corr}$ have a strong representation of the object location, and there are differences between them.
How to effectively fuse them is important.
In addition to the multiplication operation mentioned before, we believe that the summation operation can extract comprehensive features and reduce information leakage.
Therefore, we adopt a hybrid scheme to combine the cross-modal correlation features as follows:
\begin{equation}
   \begin{aligned}
      \boldsymbol{f}^{5}_{\rm clm} = (\boldsymbol{f}^{5}_{\rm r-corr} \oplus \boldsymbol{f}^{5}_{\rm t-corr} )  \circledcirc (\boldsymbol{f}^{5}_{\rm r-corr} \otimes \boldsymbol{f}^{5}_{\rm t-corr} ),
    \label{eq:2}
    \end{aligned}
\end{equation}
where $\boldsymbol{f}^{5}_{\rm clm} \in \mathbb{R}^{h_5\!\times\!w_5\!\times\!c_5}$ is the output feature of the CLM, $\oplus$/$\otimes$ and $\circledcirc$ are the element-wise multiplication/summation and concatenation, respectively, and we omit the convolutional layer.

Through the above collaborative multiplication and summation operations, we can obtain informative features of object location.
Furthermore, as shown in Fig.~\ref{CLM_structure}, we attach a location head (LocHead) after the CLM, and impose the location supervision to achieve more accurate object location.

\begin{figure}
\centering
\footnotesize
  \begin{overpic}[width=0.9\columnwidth]{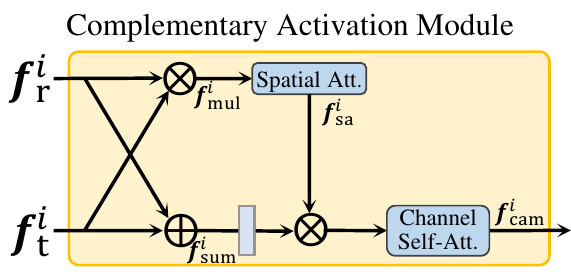}
  \end{overpic}
\caption{
Illustration of the Complementary Activation Module (CAM).
}
\label{CAM_structure}
\end{figure}

\subsection{Complementary Activation Module}
\label{sec:CAM} 
Middle-level cross-modal features (\ie $\{\boldsymbol{f}^{2\sim4}_{\rm r}, \boldsymbol{f}^{2\sim4}_{\rm t}\}$) occupy the majority and play the role of gradually refining objects and restoring the resolution.
Since we locate all potential objects in high-level features, we try to specifically activate exact regions of objects in middle-level features with different scales and propose \emph{Complementary Activation Module} to achieve it.
Concretely, our CAM is based on the attention mechanism~\cite{2018CBAM,2019DANet}, which can activate specific feature regions~\cite{2018CBAM} and establish stronger connections between features~\cite{2019DANet}.
We continue to take advantage of the two types of complementary feature combinations, \ie summation and multiplication, to mine the common valuable content.

We illustrate the CLM in Fig.~\ref{CAM_structure}, whose inputs are $\boldsymbol{f}^{i}_{\rm r}$ and $\boldsymbol{f}^{i}_{\rm t}$.
In general, the CLM contains two steps of feature combinations and feature activation.
First, we achieve feature combinations by element-wise multiplication and summation, generating $\{\boldsymbol{f}^{i}_{\rm mul}, \boldsymbol{f}^{i}_{\rm sum}\}_{i=2,3,4} \in \mathbb{R}^{h_i\!\times\!w_i\!\times\!{c}_i}$.
Since $\boldsymbol{f}^{i}_{\rm mul}$ represents the coexistence information and $\boldsymbol{f}^{i}_{\rm sum}$ represents the comprehensive information without omission, we make an attempt to enhance the broad $\boldsymbol{f}^{i}_{\rm sum}$ with the pithy $\boldsymbol{f}^{i}_{\rm mul}$ for fine region activation.
Notably, to preserve the high-response regions of $\boldsymbol{f}^{i}_{\rm mul}$, we forgo smoothing it with the convolutional layer.
We apply a convolutional layer to $\boldsymbol{f}^{i}_{\rm sum}$ for feature normalization.

Second, we apply the spatial attention~\cite{2018CBAM} to $\boldsymbol{f}^{i}_{\rm mul}$ to obtain the pithy spatial attention map, and use this map to highlight the target regions in $\boldsymbol{f}^{i}_{\rm sum}$, generating $ \boldsymbol{f}^{i}_{\rm sa} \in \mathbb{R}^{h_i\!\times\!w_i\!\times\!{c}_i}$.
In addition, we introduce the channel self-attention~\cite{2019DANet} to model the channel-wise dependencies of $\boldsymbol{f}^{i}_{\rm sa}$, generating the output feature of CAM, $\boldsymbol{f}^{i}_{\rm cam} \in \mathbb{R}^{h_i\!\times\!w_i\!\times\!c_i}$.
Using two types of attention mechanisms in spatial and channel domains can enhance the robustness of discriminative feature representations.

To sum up, we briefly formulate the above process as follows:
\begin{equation}
   \begin{aligned}
       \boldsymbol{f}^{i}_{\rm cam} = CSA \big(SA(\boldsymbol{f}^{i}_{\rm r} \otimes \boldsymbol{f}^{i}_{\rm t}) \otimes conv(\boldsymbol{f}^{i}_{\rm r} \oplus \boldsymbol{f}^{i}_{\rm t}) \big), i=2,3,4,
    \label{eq:3}
    \end{aligned}
\end{equation}
where $SA(\cdot)$ and $CSA(\cdot)$ are spatial attention and channel self-attention, respectively, and $conv(\cdot)$ is the convolutional layer.
As shown in Fig.~\ref{fig:Framework}, the CAM is arranged on three levels of cross-modal features, which is conducive to mining the multi-level complementary information and can activate the same regions of objects at different scales.

\begin{figure}
\centering
\footnotesize
  \begin{overpic}[width=1\columnwidth]{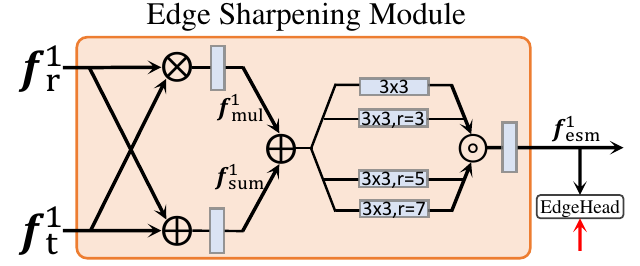}
  \end{overpic}
\caption{
Illustration of the Edge Sharpening Module (ESM).
}
\label{ESM_structure}
\end{figure}

\subsection{Edge Sharpening Module}
\label{sec:ESM}
In semantic segmentation, there are usually multiple objects of different categories in an RGB-T pair, and their sizes are varied.
In CLM and CAM, we successfully explore the common valuable content of high- and middle-level features.
Low-level features contain rich texture information, which can describe the details of objects.
Therefore, we propose \emph{Edge Sharpening Module} for low-level features to explore the common valuable content and perceive details with different receptive fields.

We illustrate the ESM in Fig.~\ref{ESM_structure}, whose inputs are $\boldsymbol{f}^{1}_{\rm r}$ and $\boldsymbol{f}^{1}_{\rm t}$.
Similar to CLM, the ESM contains feature combinations and multi-head dilated convolutions.
Feature combinations are achieved by element-wise multiplication and summation (accompanied by the convolutional layer), producing $\{\boldsymbol{f}^{1}_{\rm mul}, \boldsymbol{f}^{1}_{\rm sum}\} \in \mathbb{R}^{h_1\!\times\!w_1\!\times\!{c}_1}$.
Then, we add $\boldsymbol{f}^{1}_{\rm mul}$ and $\boldsymbol{f}^{1}_{\rm sum}$, and design a parallel structure with dilated convolutions~\cite{Dila2016} to extract multi-scale detail information.
Finally, we aggregate multi-scale features by concatenation, generating the output feature of ESM, $\boldsymbol{f}^{1}_{\rm esm} \in \mathbb{R}^{h_1\!\times\!w_1\!\times\!{c}_1}$.
We briefly summarize the above process as follows:
\begin{equation}
   \begin{aligned}
       \boldsymbol{f}^{1}_{\rm esm} = MHDC \big( conv(\boldsymbol{f}^{1}_{\rm r} \otimes \boldsymbol{f}^{1}_{\rm t}) \oplus conv(\boldsymbol{f}^{1}_{\rm r} \oplus \boldsymbol{f}^{1}_{\rm t}) \big),
    \label{eq:4}
    \end{aligned}
\end{equation}
where $MHDC(\cdot)$ represents the multi-head dilated convolutions.

As shown in Fig.~\ref{ESM_structure}, similar to CLM, we attach an edge head (EdgeHead) after the ESM, and impose the edge supervision to achieve more accurate edge information.
Moreover, considering that $\boldsymbol{f}^{1}_{\rm esm}$ is used to sharpen the edges of objects, we attach a semantic head (SemHead) after the second decoder block (\ie D$^{2}$) and impose the semantic supervision to improve the accuracy of feature representation for object regions, as shown in Fig.~\ref{fig:Framework}.
In this way, $\boldsymbol{f}^{1}_{\rm esm}$ can better sharpen the edges of object regions in the output feature of D$^{2}$.

\begin{table*}[t!]
  \centering
  \small
  \renewcommand{\arraystretch}{1.4}
  \renewcommand{\tabcolsep}{0.9mm}
  \caption{
  Quantitative comparisons (\%) on the test set of MFNet dataset. The value 0.0 represents that there are no true positives.
  `-' means that the authors do not provide the corresponding results.
  The top two results in each column are highlighted in \textcolor{red}{\textbf{red}} and \textcolor{blue}{\textbf{blue}}.
    }
\label{table:QuantitativeResults_MFNet}
  
\begin{tabular}{r|c|cccccccccccccccccc}
\midrule[1pt]    
 \multirow{2}{*}{\normalsize{Methods}} &  \multirow{2}{*}{\normalsize{Type}}
 & \multicolumn{2}{c}{Car} & \multicolumn{2}{c}{Person} & \multicolumn{2}{c}{Bike} & \multicolumn{2}{c}{Curve} & \multicolumn{2}{c}{Car Stop} & \multicolumn{2}{c}{Guardrail} & \multicolumn{2}{c}{Color Cone} & \multicolumn{2}{c}{Bump}  
 & \multirow{2}{*}{\normalsize{mAcc}} & \multirow{2}{*}{\normalsize{mIoU}} \\
 
 \cline{3-18} 
     &  & Acc & IoU & Acc & IoU & Acc & IoU & Acc & IoU & Acc & IoU & Acc & IoU & Acc & IoU & Acc & IoU \\
	     
\midrule[1pt] 
DANet$_{19}$~\cite{2019DANet}  & RGB & 89.8 & 84.5 & 65.2 & 55.0 & 76.5 & 62.6 & 44.2 & 33.4 & 32.7 & 27.4 & 2.8 & 0.9 & 46.6 & 41.9 & 56.0 & 44.5 & 57.0 & 49.7 \\	
DANet$_{19}$~\cite{2019DANet}   & RGB-T & 91.3 & 71.3 & 82.7 & 48.1 & 79.2 & 51.8 & 48.0 & 30.2 & 25.5 & 18.2 & 5.2 & 0.7 & 47.6 & 30.3 &19.9 & 18.8 & 55.2 & 41.3 \\

\hline 
HRNet$_{19}$~\cite{2019HRNet}   & RGB & 91.1 & 84.9 & 66.6 & 55.4 & 76.6 & 60.3 & 42.6 & 33.3 & 37.9 & 28.3 & 11.5 & 2.5 & 44.8 & 40.3 & 62.6 & 46.9 & 59.2 & 49.9 \\	
HRNet$_{19}$~\cite{2019HRNet}   & RGB-T & 90.8 & 86.9 & 75.1 & 67.3 & 70.2 & 59.2 & 39.1 & 35.3 & 28.0 & 23.1 & 12.1 & 1.7 & 50.4 & 46.6 & 55.8 & 47.3 & 57.9 & 51.7  \\	
								   						
\hline 
FuseNet$_{16}$~\cite{2016FuseNet}  & RGB-D & 81.0 & 75.6 & 75.2 & 66.3 & 64.5 & 51.9 & 51.0 & 37.8 & 28.7 & 15.0 & 0.0 & 0.0 & 31.1 & 21.4 & 51.9 & 45.0 & 52.4 & 45.6 \\	
D-CNN$_{18}$~\cite{2018D-CNN}     & RGB-D & 85.2 & 77.0 & 61.7 & 53.4 & 76.0 & 56.5 & 40.2 & 30.9 & 9.9 & 29.3 & 22.8 & 6.4 & 32.9 & 30.1 & 36.5 & 32.3 & 55.1 & 46.1 \\	
ACNet$_{19}$~\cite{2019ACNet}     & RGB-D & 93.7 & 79.4 & 86.8 & 64.7 & 77.8 & 52.7 & 57.2 & 32.9 & \textcolor{blue}{\textbf{51.5}} & 28.4 & 7.0 & 0.8 & 57.5 & 16.9 & 49.8 & 44.4 & 64.3 & 46.3\\	
SA-Gate$_{20}$~\cite{2020SA-Gate}    & RGB-D & 86.0 & 73.8 & 80.8 & 59.2 & 69.4 & 51.3 & 56.7 & 38.4 & 24.7 & 19.3 & 0.0 & 0.0 & 56.9 & 24.5 & 52.1 & 48.8 & 58.3 & 45.8 \\

\hline 
MFNet$_{17}$~\cite{2017MFNet}  & RGB-T & 77.2 & 65.9 & 67.0 & 58.9 & 53.9 & 42.9 & 36.2 & 29.9 & 19.1 & 9.9 & 0.1 & 8.5 & 30.3 & 25.2 & 30.0 & 27.7 & 45.1 & 39.7 \\	
RTFNet50$_{19}$~\cite{2019RTFNet}    & RGB-T & 91.3 & 86.3 & 78.2 & 67.8 & 71.5 & 58.2 & 69.8 & 43.7 & 32.1 & 24.3 & 13.4 & 3.6 & 40.4 & 26.0 & 73.5 & \textcolor{red}{\textbf{57.2}} & 62.2 & 51.7 \\	
RTFNet152$_{19}$~\cite{2019RTFNet}  & RGB-T & 93.0 & 87.4 & 79.3 & \textcolor{blue}{\textbf{70.3}} & 76.8 & 62.7 & 60.7 & \textcolor{red}{\textbf{45.3}} & 38.5 & 29.8 & 0.0 & 0.0 & 45.5 & 29.1 & \textcolor{blue}{\textbf{74.7}} & \textcolor{blue}{\textbf{55.7}} & 63.1 & 53.2 \\
PSTNet$_{20}$~\cite{2020PSTNet}  & RGB-T & - & 76.8 & - & 52.6 & - & 55.3 & - & 29.6 & - & 25.1 & - & 15.1 & - & 39.4 & - & 45.0 & - & 48.4 \\	
MLFNet$_{21}$~\cite{2021MLFNet}  & RGB-T & - & 82.3 & - & 68.1 & - & \textcolor{red}{\textbf{67.3}} & - & 27.3 & - & 30.4 & - & \textcolor{blue}{\textbf{15.7}} & - & \textcolor{red}{\textbf{55.6}} & - & 40.1 & - & 53.8 \\

FuseSeg$_{21}$~\cite{2021FuseSeg}  & RGB-T & 93.1 & \textcolor{red}{\textbf{87.9}} & 81.4 & \textcolor{red}{\textbf{71.7}} & 78.5 & \textcolor{blue}{\textbf{64.6}} & 68.4 & 44.8 & 29.1 & 22.7 & \textcolor{red}{\textbf{63.7}} & 6.4 & 55.8 & 46.9 & 66.4 & 47.9 & 70.6 & 54.5 \\	
ABMDRNet$_{21}$~\cite{2021ABMDRNet}    & RGB-T & 94.3 & 84.8 & \textcolor{red}{\textbf{90.0}} & 69.6 & 75.7 & 60.3 & 64.0 & \textcolor{blue}{\textbf{45.1}} & 44.1 & 33.1 & 31.0 & 5.1 & \textcolor{blue}{\textbf{61.7}} & 47.4 & 66.2 & 50.0 & 69.5 & \textcolor{blue}{\textbf{54.8}} \\	
MMNet$_{22}$~\cite{2022MMNet}   & RGB-T & - & 83.9 & - & 69.3 & - & 59.0 & - & 43.2 & - & 24.7 & - & 4.6 & - & 42.2 & - & 50.7 & 62.7 & 52.8 \\	
EGFNet$_{22}$~\cite{2022EGFNet}          & RGB-T & \textcolor{red}{\textbf{95.8}}& \textcolor{blue}{\textbf{87.6}} & \textcolor{blue}{\textbf{89.0}} & 69.8 & \textcolor{blue}{\textbf{80.6}} & 58.8 & \textcolor{red}{\textbf{71.5}} & 42.8 & 48.7 & \textcolor{blue}{\textbf{33.8}} & 33.6 & 7.0 & \textcolor{red}{\textbf{65.3}} & 48.3 & 71.1 & 47.1 & \textcolor{blue}{\textbf{72.7}} & \textcolor{blue}{\textbf{54.8}} \\

\hline
\hline
\textbf{LASNet (Ours)}		 & RGB-T &  \textcolor{blue}{\textbf{94.9}} & 84.2 & 81.7 & 67.1 & \textcolor{red}{\textbf{82.1}} & 56.9 & \textcolor{blue}{\textbf{70.7}} & 41.1 & \textcolor{red}{\textbf{56.8}} & \textcolor{red}{\textbf{39.6}} & \textcolor{blue}{\textbf{59.5}} & \textcolor{red}{\textbf{18.9}} & 58.1& \textcolor{blue}{\textbf{48.8}} & \textcolor{red}{\textbf{77.2}} & 40.1 & \textcolor{red}{\textbf{75.4}} & \textcolor{red}{\textbf{54.9}}    \\			   
\toprule[1pt]
\end{tabular}
\end{table*}

\subsection{Loss Function}
\label{sec:Loss Function}
As shown in Fig.~\ref{fig:Framework}, in addition to the semantic supervision on D$^{2}$, the decoder also has a semantic supervision on the final segmentation result $\mathbf{S}_{sem}$.
Therefore, during the training phase, our LASNet has a total of four supervisions, including a location supervision, an edge supervision, and two semantic supervisions.
Specifically, for the location supervision (${L}_{ loc}$), the edge supervision (${L}_{eg}$) and the semantic supervision on D$^{2}$ (${L}_{ sem2}$), we adopt the widely used weighted cross-entropy loss.
For the final semantic supervision (${L}_{ sem}$), we adopt a hybrid loss, including the weighted cross-entropy loss and the Lovasz-softmax loss~\cite{Lovaszloss}.
To make the network more focused on the semantic segmentation result, we reduce the attention of the location supervision by empirically setting a coefficient of 0.5 for it.
Compared to the semantic supervision, the edge supervision has a small loss that does not affect the semantic segmentation result, so we do not adjust it.

Here, we summarize the total training loss function ${L}_{ total}$ as follows:
\begin{equation}
   \begin{aligned}
   \left\{
	\begin{array}{lll}
	    {L}_{ total}    =    0.5{L}_{ loc} + {L}_{ eg} + {L}_{ sem2} + {L}_{ sem}, \\
	    {L}_{ loc/eg/sem2}    =   \ell_{wbce}( up(\mathbf{S}_{ loc/eg/sem2}),  \mathbf{GT}_{ loc/eg/sem}), \\
	    {L}_{ sem}     =   \ell_{wbce}( \mathbf{S}_{ sem},  \mathbf{GT}_{ sem}) + \ell_{Lovasz}( \mathbf{S}_{ sem},  \mathbf{GT}_{ sem}),\\
	\end{array}  \right. 
    \label{eq:5}
    \end{aligned}
\end{equation}
where $\mathbf{S}_{ loc}$/$\mathbf{S}_{eg}$ is generated by the LocHead/EdgeHead, $\mathbf{S}_{sem2}$ and $\mathbf{S}_{sem}$ are generated by the SemHead, and $\mathbf{GT}$ is the corresponding ground truth.
We implement the above predicted heads using a convolutional layer.
Notably, existing RGB-T semantic segmentation datasets only provide the ground truth of semantic segmentation, so we make $\mathbf{GT}_{loc}$ and $\mathbf{GT}_{eg}$ algorithmically.
We assign the background class of $\mathbf{GT}_{sem}$ as background with value of zero and the other classes as foreground with value of one, obtaining the binary location ground truth $\mathbf{GT}_{loc}$.
We extract the edges of all classes except the background class, and set the pixel values in the edges to one and others to zero, composing the binary edge ground truth $\mathbf{GT}_{eg}$.
With the effective training loss, our LASNet can converge well and produce satisfactory segmentation results.

\section{Experiments}
\label{sec:exp}

\subsection{Experimental Protocol}
\label{sec:ExpProtocol}
\textit{1) Datasets.}
We train and evaluate the proposed method on MFNet dataset~\cite{2017MFNet} and PST900 dataset~\cite{2020PSTNet}.

\textbf{MFNet dataset}\footnote{https://www.mi.t.u-tokyo.ac.jp/static/projects/mil\_multispectral/} contains 1,569 RGB-T image pairs captured by the InfReC R500 camera and corresponding pixel-level semantic annotations. 
820 RGB-T image pairs are taken at daytime, and the remaining 749 pairs are at nighttime.
The resolution of all RGB and TIR images is 480$\times$640.
In addition to the background class, this dataset contains eight classes, including car, person, bike, curve, car stop, guardrail, color cone, and bump.
All RGB-T image pairs are divided into three parts, of which 784 pairs (410 daytime images and 374 nighttime images) are used for training, 392 pairs (205 daytime images and 187 nighttime images) for validation, and 393 pairs (205 daytime images and 188 nighttime images) for testing.

\textbf{PST900 dataset}\footnote{https://github.com/ShreyasSkandanS/pst900\_thermal\_rgb} contains 894 aligned RGB and TIR image pairs, where RGB images are captured by a Stereolabs ZED Mini stereo camera and TIR images are captured by a FLIR Boson 320 camera.
The resolution of all RGB and TIR images is 1280$\times$720.
This dataset contains four classes of hand-drill, backpack, fire-extinguisher and survivor, and a default background class.
This dataset is divided into two parts for training and testing, of which 597 pairs\footnote{Notably, there are actually only 597 pairs in the training set of previous works~\cite{2020PSTNet,2022MFFENet,2022EGFNet,2022MTANet} rather than 606 pairs.} constitute the training set and 288 pairs constitute the testing set.

\textit{2) Evaluation Metrics.}
We adopt two widely used quantitative evaluation metrics to evaluate the segmenattion performance of our method and all compared methods, including mean accuracy (mAcc) and mean intersection over union (mIoU), which can be respectively calculated as follows:
\begin{equation}
   \begin{aligned}
	mAcc = \frac{1}{C} \sum\limits_{k=1}^C \frac{TP_{k}}{TP_{k}+FN_{k}},
    \label{eq:6}
    \end{aligned}
\end{equation}
\begin{equation}
   \begin{aligned}
	mIoU = \frac{1}{C} \sum\limits_{k=1}^C \frac{TP_{k}}{TP_{k}+FP_{k}+FN_{k}},
    \label{eq:6}
    \end{aligned}
\end{equation}
where $C$ is the total number of classes including background class, which is respectively set to nine and five in the MFNet dataset and PST900 dataset, and $TP_{k}$, $FP_{k}$ and $FN_{k}$ represent the true positives, false positives and false negatives of class $k$, respectively.
The higher the two metrics, the better.

\textit{3) Implementation Details.}
We implement the proposed LASNet on the PyTorch~\cite{PyTorch} with an NVIDIA GTX 3090 GPU (24GB RAM).
For the MFNet dataset and PST900 dataset, we train and test the proposed LASNet according to the dataset partition described earlier.
During the training phase, we keep the original resolution of RGB image and TIR image for network input, use the random flipping and cropping for data augmentation, and adopt the Ranger optimizer with the batch size of 4 and the initial learning rate of $5e^{-5}$.
The parameters of the feature extractor are initialized by the pre-trained ResNet-152 model~\cite{2016ResNet}, and the parameters of other newly added convolutional layers are initialized through the ``Kaiming" method~\cite{InitialWei}.
For both datasets, we train the proposed LASNet for 200 epochs.
During the testing phase, we directly input the RGB-T image pair with original resolution into the trained LASNet without any post-processing to obtain the segmentation result.

\begin{table}[t!]
  \centering
  \small
  \renewcommand{\arraystretch}{1.4}
  \renewcommand{\tabcolsep}{2.6mm}
  \caption{
    Quantitative comparisons (\%) on the test set of MFNet dataset in daytime and nighttime.
   The best result in each column is highlighted in \textcolor{red}{\textbf{red}}.
    }
\label{table:MFNet_DayNight}
  
\begin{tabular}{r|c|cccc}
\midrule[1pt]    
 \multirow{2}{*}{\normalsize{Methods}}& \multirow{2}{*}{\normalsize{Type}}
 & \multicolumn{2}{c}{Daytime} & \multicolumn{2}{c}{Nighttime} \\
 
 \cline{3-6} 
       & & Acc & IoU & Acc & IoU \\
	     
\midrule[1pt] 
DANet$_{19}$~\cite{2019DANet}   & RGB & 61.0 & 46.3 & 52.6 & 47.0  \\
DANet$_{19}$~\cite{2019DANet}   & RGB-T & 50.9 & 37.5 & 52.4 & 40.1  \\
\hline
HRNet$_{19}$~\cite{2019HRNet}   & RGB & 64.7 & 46.7 & 54.0 & 47.3  \\
HRNet$_{19}$~\cite{2019HRNet}   & RGB-T & 54.4 & 46.1 & 55.1 & 50.7  \\								   						
							   						
\hline 
FuseNet$_{16}$~\cite{2016FuseNet}   & RGB-D & 49.5 & 41.0 & 48.9 & 43.9  \\
D-CNN$_{18}$~\cite{2018D-CNN}     & RGB-D & 50.6 & 42.4 & 50.7 & 43.2  \\
ACNet$_{19}$~\cite{2019ACNet}     & RGB-D & 60.7 & 41.6 & 63.9 & 47.4  \\
SA-Gate$_{20}$~\cite{2020SA-Gate}   & RGB-D & 49.3 & 37.9 & 56.9 & 45.6  \\
							   						
\hline 
MFNet$_{17}$~\cite{2017MFNet}          & RGB-T & 42.6 & 36.1 & 48.9 & 43.9  \\
RTFNet50$_{19}$~\cite{2019RTFNet}    & RGB-T & 57.3 & 44.4 & 59.4 & 52.0  \\	
RTFNet152$_{19}$~\cite{2019RTFNet}  & RGB-T & 60.0 & 45.8 & 60.7 & 54.8  \\
MLFNet$_{21}$~\cite{2021MLFNet}   & RGB-T & - & 45.6 & - & 54.9  \\

FuseSeg$_{21}$~\cite{2021FuseSeg}  & RGB-T & 62.1 & \textcolor{red}{\textbf{47.8}} & 67.3 & 54.6  \\
ABMDRNet$_{21}$~\cite{2021ABMDRNet}    & RGB-T & 58.4 & 46.7 & 68.3 & 55.5 \\

\hline
\hline
\textbf{LASNet (Ours)} & RGB-T & \textcolor{red}{\textbf{73.3}} & 45.2 & \textcolor{red}{\textbf{72.8}} & \textcolor{red}{\textbf{58.7}} \\	   
\toprule[1pt]
\end{tabular}
\end{table}
\begin{figure*}[t!]
    \centering
    \footnotesize
	\begin{overpic}[width=1\textwidth]{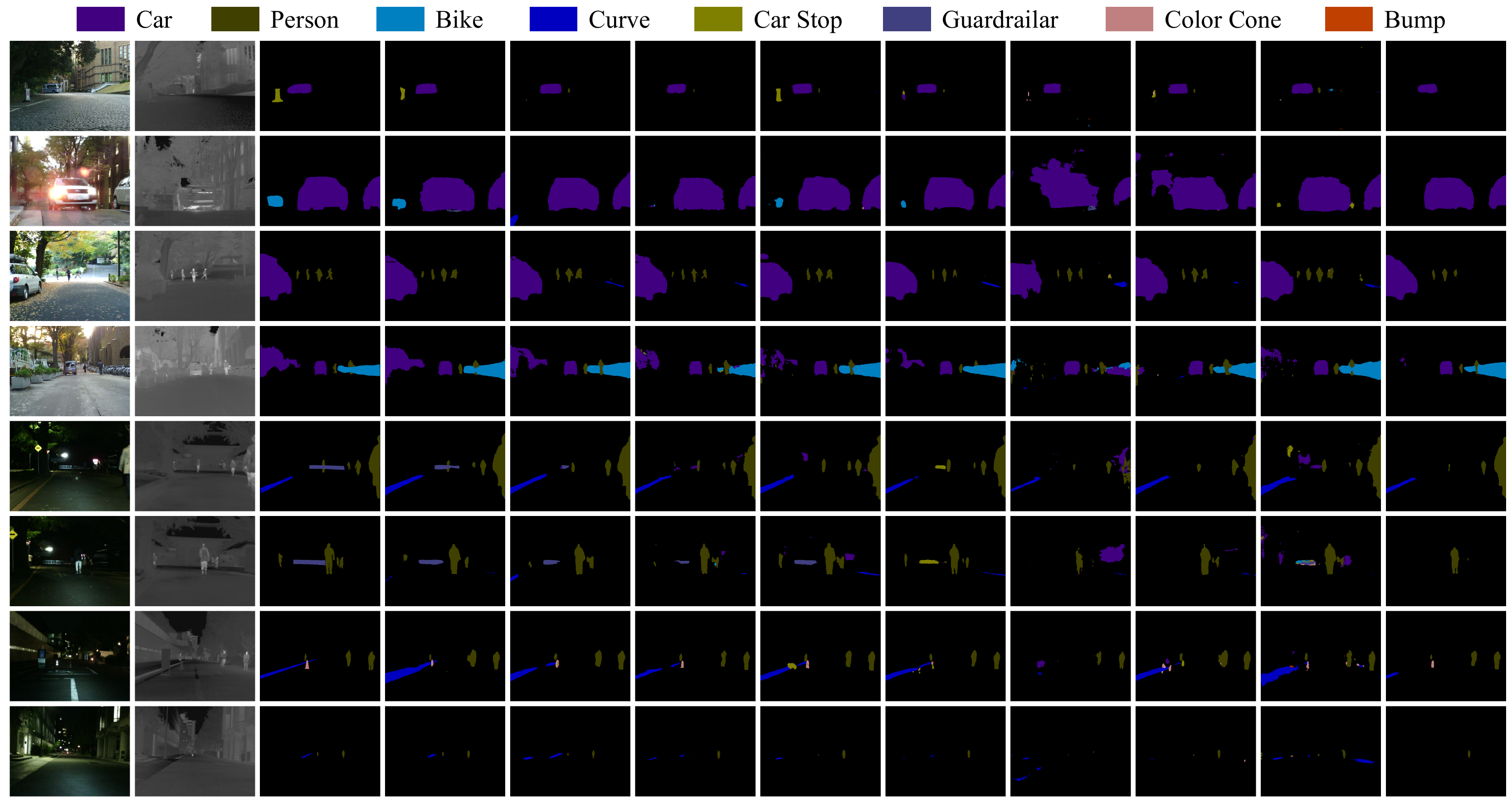}

    \put(2.42,-1.1){ RGB }
    \put(11.1,-1.1){ TIR}
    \put(19.57,-1.1){ GT }
    \put(27.2,-1.1){ \textbf{Ours} }
    \put(34.58,-1.1){ EGFNet }
    \put(42.98,-1.1){ MMNet }
    \put(49.73,-1.1){ ABMDRNet }
    \put(59.5,-1.1){ RTFNet }
    \put(68.5,-1.1){MFNet }
    \put(75.8,-1.1){ SA-Gate }
    \put(84.45,-1.1){ ACNet }
    \put(92.1,-1.1){ HRNet-T } 
    
    \end{overpic}
	\caption{Visual comparisons of our method and eight representative state-of-the-art methods in daytime (the first four cases) and nighttime (the last four cases) on the test set of MFNet.
	HRNet-T represents the modified HRNet for RGB-T image pairs.
	Please zoom-in for the best view.
    }
    \label{fig:VisualExample}
\end{figure*}
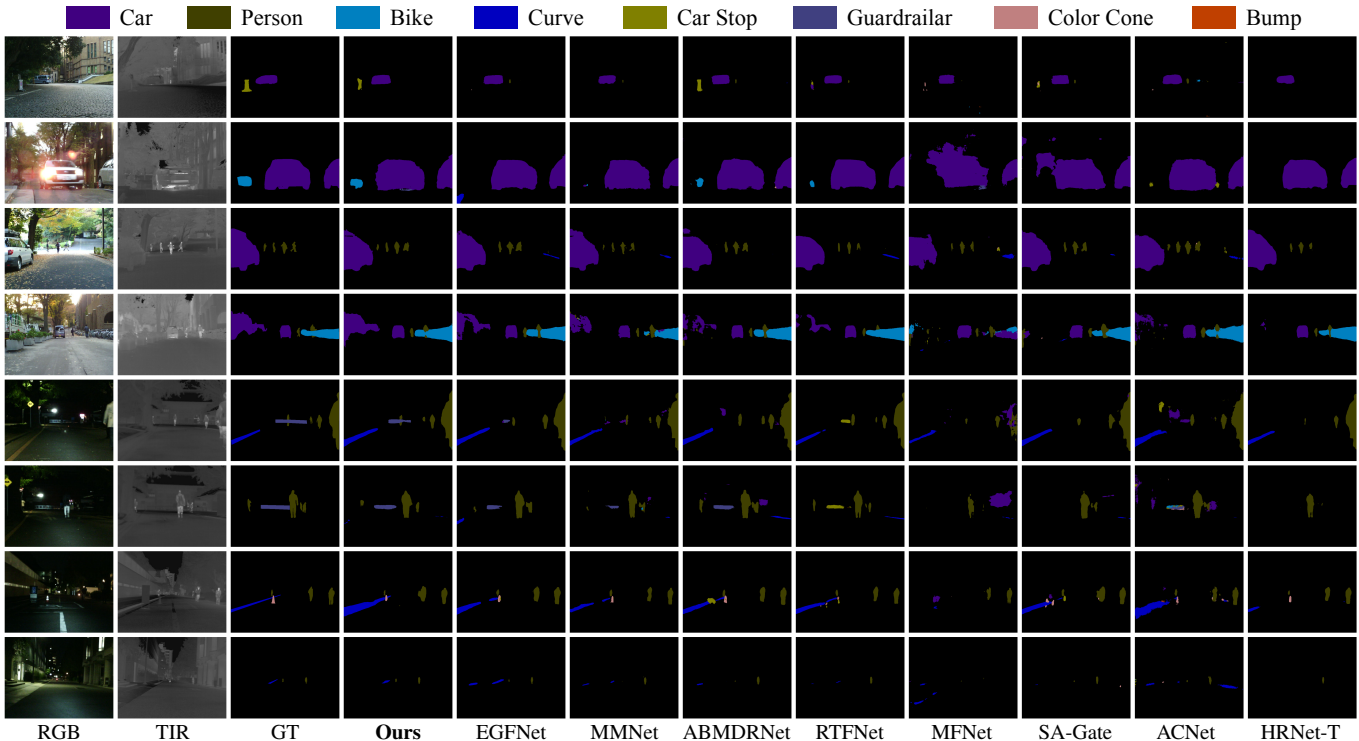

\subsection{Comparison with State-of-the-arts}
We compare our LASNet with relevant state-of-the-art RGB/RGB-D/RGB-T semantic segmentation methods.
For a fair comparison, following previous works~\cite{2021ABMDRNet,2022EGFNet}, we also modify some RGB semantic segmentation methods to adapt to RGB-T image pairs.
We retrain the RGB, RGB-D, and modified RGB methods with their default parameter settings on the same training set as our method on two datasets.
We obtain the segmentation maps of other RGB-T methods through the public benchmarks or codes.

\textit{1) Comparison on the MFNet Dataset.}
On the MFNet dataset, we compare our LASNet with 14 state-of-the-art methods,
including two RGB semantic segmentation methods (\ie DANet~\cite{2019DANet} and HRNet~\cite{2019HRNet}) and their modified RGB-T versions,
four RGB-D semantic segmentation methods (\ie FuseNet~\cite{2016FuseNet}, D-CNN~\cite{2018D-CNN}, ACNet~\cite{2019ACNet}, and SA-Gate~\cite{2020SA-Gate}),
and eight RGB-T semantic segmentation methods (\ie MFNet~\cite{2017MFNet}, two versions of RTFNet~\cite{2019RTFNet}, PSTNet~\cite{2020PSTNet}, MLFNet~\cite{2021MLFNet}, FuseSeg~\cite{2021FuseSeg}, ABMDRNet~\cite{2021ABMDRNet}, MMNet~\cite{2022MMNet}, and EGFNet~\cite{2022EGFNet}).

We report the quantitative performance of our method and all compared methods on the MFNet dataset in Tab.~\ref{table:QuantitativeResults_MFNet}, including performance of eight classes and overall performance.
Overall, our method achieves the best performance in terms of two metrics, especially on mAcc.
Across all 18 metrics, our method achieves seven first places and four second places, demonstrating a competitive adaptability to different scenes.
Concretely, on mAcc, our method outperforms the second best method (\ie EGFNet) by a large margin, reaching 2.7\%.
On mIoU, our method outperforms the second best method (\ie EGFNet and ABMDRNet) by 0.1\%.
Among all classes, our method performs very well on the class of car stop, surpassing EGFNet by 8.1\% and 5.8\% on mAcc and mIoU, respectively.

In addition, we report the quantitative performance of some available methods on the test set of MFNet dataset in daytime and nighttime in Tab.~\ref{table:MFNet_DayNight}.
Our method shows its superiority in both scenes, especially in the nighttime scene.
This means that the three steps (\ie object location, region activation, and edge sharpening) of our method is successful and can make full use of TIR images to handle low illumination scenes.
The above quantitative analysis clearly demonstrates the effectiveness of our LASNet on the MFNet dataset.

\begin{table*}[t!]
  \centering
  \small
  \renewcommand{\arraystretch}{1.4}
  \renewcommand{\tabcolsep}{1.2mm}
  \caption{
    Quantitative comparisons (\%) on the test set of PST900 dataset.
    `-' means that the authors do not provide the corresponding results.
    The top two results in each column are highlighted in \textcolor{red}{\textbf{red}} and \textcolor{blue}{\textbf{blue}}.
    }
\label{table:QuantitativeResults_PST900}
  
\begin{tabular}{r|c|cccccccccccc}
\midrule[1pt]    
 \multirow{2}{*}{\normalsize{Methods}} &  \multirow{2}{*}{\normalsize{Type}}
 & \multicolumn{2}{c}{Background} & \multicolumn{2}{c}{Hand-Drill} & \multicolumn{2}{c}{Backpack} & \multicolumn{2}{c}{Fire-Extinguisher} & \multicolumn{2}{c}{Survivor}   
 & \multirow{2}{*}{\normalsize{mAcc}} & \multirow{2}{*}{\normalsize{mIoU}} \\
 
 \cline{3-12} 
       & & Acc & IoU & Acc & IoU & Acc & IoU & Acc & IoU & Acc & IoU \\
	     
\midrule[1pt] 

ERFNet$_{18}$~\cite{2018ERFNet}   & RGB & - & 98.69 & - & 42.40 & - & 65.28 & - & 61.18 & - & 41.69 & - & 61.85 \\
ERFNet$_{18}$~\cite{2018ERFNet}   & RGB-T & - & 98.73 & - & 52.76 & - & 68.08 & - & 58.79 & - & 34.38 & - & 62.55 \\
\hline
CCNet$_{19}$~\cite{2019CCNet}   & RGB & \textcolor{red}{\textbf{99.86}} & 99.05 & 51.77 & 32.27 & 68.30 & 66.42 & 67.79 & 51.84 & 60.84 & 57.50 & 69.71 & 61.42 \\
CCNet$_{19}$~\cite{2019CCNet}   & RGB-T & 99.59 & 98.74 & 54.09 & 51.01 & 75.96 & 72.95 & 88.06 & \textcolor{blue}{\textbf{73.80}} & 49.45 & 33.52 & 73.43 & 66.00 \\
\hline
EfficientFCN$_{20}$~\cite{2020EfficientFCN}  & RGB & 99.81 & 98.63 & 32.08 & 30.12 & 60.06 & 58.15 & 78.87 & 39.96 & 32.76 & 28.00 & 60.72 & 50.98 \\	
EfficientFCN$_{20}$~\cite{2020EfficientFCN}  & RGB-T & 99.80 & 98.85 & 48.75 & 38.58 & 69.90 & 67.59 & 76.45 & 46.28 & 38.86 & 35.06 & 66.75 & 57.27 \\	
								   						
\hline 
ACNet$_{19}$~\cite{2019ACNet}       & RGB-D & \textcolor{blue}{\textbf{99.83}}& 99.25 & 53.59 & 51.46 & 85.56 & 83.19 & 84.88 & 59.95 & 69.10 & 65.19 & 78.67 & 71.81 \\	
SA-Gate$_{20}$~\cite{2020SA-Gate}    & RGB-D & 99.74 & 99.25 & 89.88 & \textcolor{red}{\textbf{81.01}} & 89.03 & 79.77 & 80.70 & 72.97 & 64.19 & 62.22 & 84.71 & \textcolor{blue}{\textbf{79.05}} \\	
							   						
\hline 
MFNet$_{17}$~\cite{2017MFNet}            & RGB-T & - & 98.63 & - & 41.13 & - & 64.27 & - & 60.35 & - & 20.70 & - & 57.02 \\	
PSTNet$_{20}$~\cite{2020PSTNet}           & RGB-T & - & 98.85 & - & 53.60 & - & 69.20 & - & 70.12 & - & 50.03 & - & 68.36 \\	
MFFENet$_{22}$~\cite{2022MFFENet}       & RGB-T & - & \textcolor{blue}{\textbf{99.40}} & - & 72.50 & - & 81.02 & - & 66.38 & - & \textcolor{blue}{\textbf{75.60}} & - & 78.98 \\	
EGFNet$_{22}$~\cite{2022EGFNet}           & RGB-T & 99.48 & 99.26 & \textcolor{red}{\textbf{97.99}} & 64.67 & \textcolor{red}{\textbf{94.17}} & 83.05 & \textcolor{red}{\textbf{95.17}} & 71.29 & \textcolor{blue}{\textbf{83.30}} & 74.30 & \textcolor{red}{\textbf{94.02}} & 78.51 \\
MTANet$_{22}$~\cite{2022MTANet}            & RGB-T & - & 99.33 & - & 62.05 & - & \textcolor{red}{\textbf{87.50}} & - & 64.95 & - & \textcolor{red}{\textbf{79.14}} & - & 78.60 \\

\hline
\hline
\textbf{LASNet (Ours)}		 & RGB-T & 99.77 & \textcolor{red}{\textbf{99.46}} & \textcolor{blue}{\textbf{92.36}} & \textcolor{blue}{\textbf{77.75}} & \textcolor{blue}{\textbf{90.80}} & \textcolor{blue}{\textbf{86.48}} & \textcolor{blue}{\textbf{91.81}} & \textcolor{red}{\textbf{82.80}} & \textcolor{red}{\textbf{83.43}} & 75.49 & \textcolor{blue}{\textbf{91.63}} & \textcolor{red}{\textbf{84.40}} \\
\toprule[1pt]
\end{tabular}
\end{table*}
Furthermore, to visually compare the segmentation results, we show the segmentation results of our method and eight representative state-of-the-art methods in daytime and nighttime on the test set of MFNet in Fig.~\ref{fig:VisualExample}.
The the first four cases belong to daytime, and the the last four belong to nighttime.
In these four daytime cases with cluttered backgrounds, our method can accurately identify and locate all categories of objects in the first three cases, while other methods miss or mis-segment objects.
This benefits from the CLM and CAM in our method, where the CLM is responsible for accurate location and the CAM is responsible for exact region activation.
In the last case of the daytime scene, the objects segmented by our method are more complete than those of other methods, which benefits from the precise activation of CAM and the edge sharpening of ESM.
In these four nighttime cases with adverse illumination conditions, it is obvious that the objects are mostly barely visible in the RGB images, but clearly in the TIR images.
Thanks to the valuable information mined by the three proposed modules from cross-modal features, our method achieves satisfactory segmentation results in all difficult nighttime scenes even in small object scenes (the last case).
In summary, our method can handle both challenging scenes to generate good segmentation results, showing strong generalization.

\textit{2) Comparison on the PST900 Dataset.}
On the PST900 dataset, we compare our LASNet with 10 state-of-the-art methods,
including three RGB semantic segmentation methods (\ie ERFNet~\cite{2018ERFNet}, CCNet~\cite{2019CCNet}, and EfficientFCN~\cite{2020EfficientFCN}) and their modified RGB-T versions,
two RGB-D semantic segmentation methods (\ie ACNet~\cite{2019ACNet} and SA-Gate~\cite{2020SA-Gate}),
and five RGB-T semantic segmentation methods (\ie MFNet~\cite{2017MFNet}, PSTNet~\cite{2020PSTNet}, MFFENet~\cite{2022MFFENet}, EGFNet~\cite{2022EGFNet}, and MTANet~\cite{2022MTANet}).

We report the quantitative performance of our method and all compared methods on the PST900 dataset in Tab.~\ref{table:QuantitativeResults_PST900}, including performance of five classes and overall performance.
Overall, our method achieves competitive performance on the PST900 dataset.
Across all 12 metrics, our method achieves four first places and six second places.
Specifically, our method ranks first on mIoU, outperforming the second-place SA-Gate by 5.35\%.
Our method ranks second on mAcc, 2.46\% lower than EGFNet, but 5.89\% ahead of EGFNet on mIoU.
Notably, our method improves the IoU score for the fire-extinguisher class by an astonishing 9.00\%.
The above analysis demonstrates the effectiveness of our LASNet on the PST900 dataset, as well as its applicability on different datasets.

\begin{table}[!t]
\centering
\small
\caption{Quantitative results (\%) of assessing the individual and joint contributions of the three modules in LASNet.
  The best one is \textcolor{red}{\textbf{red}}.
  }
\label{Ablation_component}
\renewcommand{\arraystretch}{1.45}
\renewcommand{\tabcolsep}{2.8mm}
\begin{tabular}{c|cccc||c}
\bottomrule

 \multirow{2}{*}{No.} & \multirow{2}{*}{Baseline} & \multirow{2}{*}{ESM} & \multirow{2}{*}{\rm CAM} & \multirow{2}{*}{CLM}  
 & MFNet~\cite{2017MFNet} \\
 
 \cline{6-6}
    & & & & 
    & mIoU  \\
\hline
\hline
1 &  \Checkmark &                      &                      &                     & 48.3   \\ 
2 &  \Checkmark & \Checkmark  &                      &                     & 50.9   \\
3 &  \Checkmark &                      & \Checkmark &                      & 50.5   \\
4 &  \Checkmark &                      &                     & \Checkmark  & 50.4  \\
\hline
5 &  \Checkmark &                      & \Checkmark & \Checkmark  & 52.6   \\
6 &  \Checkmark & \Checkmark  &                     & \Checkmark  & 52.0   \\
7 &  \Checkmark & \Checkmark  & \Checkmark &                      & 52.0   \\

\hline
8 &  \Checkmark & \Checkmark  & \Checkmark & \Checkmark  & \textcolor{red}{\textbf{54.9}}  \\
\toprule
\end{tabular}
\end{table}

\subsection{Ablation Studies}
\label{Ablation Studies}
We conduct comprehensive ablation experiments to evaluate the effectiveness of each module of our LASNet on the MFNet dataset.
Specifically, we evaluate
the individual and joint contributions of the three modules,
the effectiveness of each component of ESM, CAM and CLM, and
the validity of auxiliary location supervision and edge supervision.
For all ablation experiments, we train the variant with the same parameter and dataset settings as described in Sec.~\ref{sec:ExpProtocol}, and adopt mIoU to evaluate the performance.

\textit{1) The individual and joint contributions of the three modules}.
We propose ESM, CAM and CLM to achieve three steps of location, activation and sharpening with discriminative feature processing.
Here, we provide four variants to assess the individual contribution of the three modules:
1) Baseline, where we employ element-wise summation to fuse RGB and TIR features instead of three modules,
2) Baseline+ESM, 
3) Baseline+CAM, and
4) Baseline+CLM.
The quantitative results are reported in Tab.~\ref{Ablation_component}.
``Baseline" only achieves 48.3\% on mIoU, which is 6.6\% lower than our full LASNet, demonstrating that the three modules can indeed improve the segmentation accuracy.
With the help of ESM, CAM or CLM, No.2, No.3, and No.4 variants improve the performance compared to ``Baseline", respectively.

Furthermore, we provide three variants to assess the joint contribution of the three modules in Tab.~\ref{Ablation_component}:
5) Baseline+CAM+CLM, 
6) Baseline+ESM+CLM, and
7) Baseline+ESM+CAM.
With the cooperation of two modules, the performance of the above three variants is further improved compared to that of a single module.
The perfect cooperation of the three modules makes the excellent full LASNet, reaching 54.9\%.
The above analysis shows that the proposed modules are valid and contribute to the final segmentation performance.

\begin{table}[!t]
\centering
\small
\caption{Quantitative results (\%) of assessing the effectiveness of each component of ESM, CAM, and CLM.
  The best one is \textcolor{red}{\textbf{red}}.
  }
\label{Ab_ESMCAMCLM}
\renewcommand{\arraystretch}{1.4}
\renewcommand{\tabcolsep}{4.3mm}
\small
\begin{tabular}{c|c||c}
\bottomrule
   \multirow{2}{*}{Aspects} & \multirow{2}{*}{Models}  & MFNet~\cite{2017MFNet}  \\
 \cline{3-3}
  &   & mIoU  \\
\hline
\hline
& \textbf{LASNet (Ours)}  & \textcolor{red}{\textbf{54.9}}  \\

\hline
  \multirow{3}{*}{\begin{sideways}ESM\end{sideways}}
& \textit{w/o mul}       	       &  53.3   \\ 
& \textit{w/o sum}        	       &  53.2   \\ 
& \textit{w/o MHDC}                 & 53.5    \\ 

\hline
  \multirow{3}{*}{\begin{sideways}CAM\end{sideways}}
& \textit{w/o mul2}              &  53.7   \\
& \textit{w/o SA}               &  52.7 \\ 
& \textit{w/o CSA}            &  52.9 \\

\hline
  \multirow{3}{*}{\begin{sideways}CLM\end{sideways}}
& \textit{w/o Corr}         & 52.3     \\ 
& \textit{w/o mul3}         &  53.2  \\ 
& \textit{w/o sum3}       &  53.0   \\ 

\toprule
\end{tabular}
\end{table}

\textit{2) The effectiveness of each component of ESM.}
To assess the effectiveness of each component of ESM, we provide three variants of ESM in the upper part of Tab.~\ref{Ab_ESMCAMCLM}:
1) removing the feature combination of multiplication, \ie \textit{w/o mul},
2) removing the feature combination of summation, \ie \textit{w/o sum}, and 
3) replacing multi-head dilated convolutions with a convolutional layer, \ie \textit{w/o MHDC}.
We observe that mining multi-scale details information in only one feature combination (multiplication or summation) is unsatisfactory, resulting in performance degradation.
The similar performance of \textit{w/o mul} and \textit{w/o sum}, \ie 53.3\% and 53.2\%, indicates these two types of feature combinations are equally important in ESM.
\textit{w/o MHDC} means that only the single-scale detail information can be extracted, which is not conducive to detecting the edges of objects of different scales.
Therefore, the performance of \textit{w/o MHDC} drops.

\textit{3) The effectiveness of each component of CAM.}
In CAM, we enhance $\boldsymbol{f}^{i}_{\rm sum}$ with $\boldsymbol{f}^{i}_{\rm mul}$ for fine region activation.
To assess the effectiveness of this process, we provide a variant, which removes the feature combination of multiplication and employs self-enhancement on $\boldsymbol{f}^{i}_{\rm sum}$, \ie \textit{w/o mul2}.
The results shown in Tab.~\ref{Ab_ESMCAMCLM} illustrate that the self-enhancement of $\boldsymbol{f}^{i}_{\rm sum}$ is sub-optimal, and that the multiplication operation can extract more compact information that is conducive to accurate region activation.
Besides, we remove the spatial attention, \ie \textit{w/o SA}, to prove the effectiveness of the pithy spatial attention map in region activation.
Moreover, we remove the channel self-attention, \ie \textit{w/o CSA}, to illustrate the importance of establishing channel-wise dependencies for region activation.
The dropped performance of \textit{w/o SA} and \textit{w/o CSA} in Tab.~\ref{Ab_ESMCAMCLM} means that using one kind of attention alone is insufficient, and that combining the attention of two different domains is powerful for exploring discriminative information.

\textit{4) The effectiveness of each component of CLM.}
To assess the effectiveness of each component of CLM, we provide three variants of CLM in the bottom part of Tab.~\ref{Ab_ESMCAMCLM}:
1) removing the cross-modal correlation modeling, \ie \textit{w/o Corr},
2) removing the correlation combination of multiplication, \ie \textit{w/o mul3}, and 
3) removing the correlation combination of summation, \ie \textit{w/o sum3}. 
\textit{w/o Corr} discards building cross-modal pixel-level correlations that can collaboratively locate objects, so its performance drops by 2.6\%, which is the worst in Tab.~\ref{Ab_ESMCAMCLM}.
This indicates that the cross-modal correlation model is extremely important.
It is sub-optimal to use only one combination operation to fuse cross-modal correlations, resulting in a 1.7\% performance drop for \textit{w/o mul3} and a 1.9\% performance drop for \textit{w/o sum3}. 
This phenomenon is the same as the performance degradation of \textit{w/o mul} and \textit{w/o sum} of ESM, which justifies the rationality of our LASNet in exploring the common valuable content of cross-modal features at different levels through multiplication and summation.

\begin{table}[!t]
\centering
\small
\caption{Quantitative results (\%) of illustrating the advantages of the combination of the element-wise multiplication and the element-wise summation in our three modules.
M means the element-wise multiplication, S means the element-wise summation, and C means the concatenation.
  The best one is \textcolor{red}{\textbf{red}}.
  }
\label{Ablation_MulSum}
\renewcommand{\arraystretch}{1.4}
\renewcommand{\tabcolsep}{3.3mm}
\begin{tabular}{c||ccc}
\bottomrule

Models & \textbf{M+S (Ours)} & C+S & M+C  \\%

\hline

mIoU  & \textcolor{red}{\textbf{54.9}} & 54.2  & 53.6  \\

\hline
\hline

Models & S+M & S+C  & C+M \\%

\hline

mIoU  & 53.3 & 53.6 & 53.7  \\

\toprule

\end{tabular}
\end{table}

\textit{5) The advantages of the combination of the element-wise multiplication and the element-wise summation in our three modules.}
To assess the advantages of the combination of the element-wise multiplication and the element-wise summation in our three modules, we provide five variants for feature integration in Tab.~\ref{Ablation_MulSum} (M, S, and C means the element-wise multiplication, the element-wise summation, and the concatenation, respectively):
1) C+S means replacing M with C and keeping S,
2) M+C,
3) S+M,
4) S+C, and
5) C+M.
Here, we define our original structure as M+S.
We observe that the performance of all five variants is inferior to our M+S, which illustrates the optimality of our combination of M and S.
We believe that the better performance of M+S can be attributed to
1) the reasonable use of the coexistence information mined by M and the comprehensive information mined by S, and
2) successful extraction of complementary information of the above information for segmentation.
Besides, the reasons for the poor performance of other variants may be as follows:
1) the concatenation is a general operation for feature integration, but its generated features cannot effectively express the complementary information of the cross-model features, and
2) the position of operations affects the extraction of complementary information, and the exchange of position will lead to the failure to extract effective complementary information.

\begin{table}[!t]
\centering
\small
\caption{Quantitative results (\%) of assessing the validity of auxiliary location supervision (LocSup) and edge supervision (EdgeSup).
The number in parentheses is the coefficient of LocSup.
  The best one is \textcolor{red}{\textbf{red}}.
  }
\label{Ablation_LocEG}
\renewcommand{\arraystretch}{1.4}
\renewcommand{\tabcolsep}{2.2mm}
\begin{tabular}{c|cc||c}
\bottomrule

 \multirow{2}{*}{No.} & \multirow{2}{*}{LocSup} & \multirow{2}{*}{EdgeSup}  
 & MFNet~\cite{2017MFNet}  \\
 
 \cline{4-4}
    & & & mIoU  \\
\hline
\hline
1 &                               &                      &   53.0 \\
2 &  \Checkmark  (0.5) &                      &   53.9 \\
3 &                                & \Checkmark  &   53.4 \\
4 &  \Checkmark (1)     &  \Checkmark  &  53.9 \\

\hline
5 &  \Checkmark (0.5) & \Checkmark  & \textcolor{red}{\textbf{54.9}}  \\
\toprule
\end{tabular}
\end{table}

\textit{6) The validity of auxiliary location supervision and edge supervision.}
It is important to employ proper supervision in the training phase, which can improve the segmentation accuracy without increasing the computational cost in the inference phase.
To assess the validity of auxiliary location supervision and edge supervision attached to CLM and ESM, we provide three combinations of supervision with default coefficients in Tab.~\ref{Ablation_LocEG}:
1) removing both supervisions,
2) keeping only location supervision, and
3) keeping only edge supervision.
We observe that discarding one or all of the supervision is not good for the final segmentation performance.
Under the effective guidance of both supervisions, CLM and ESM can enhance the representation of object location and fine edges, contributing to the final segmentation results of our LASNet.
Moreover, to prove the rationality of coefficient setting of location supervision, we provide a variant of the location supervision with a coefficient of 1.
As shown in Tab.~\ref{Ablation_LocEG}, we observe a 1.0\% decrease in the performance of the last variant, which means that it is reasonable to reduce the weight of location supervision.

\section{Conclusion}
\label{sec:con}
In this paper, we follow the feature fusion paradigm, and propose \emph{LASNet} for RGB-T semantic segmentation.
The main idea of LASNet is to process RGB and TIR features at different levels with specific modules.
Therefore, we propose three plug-and-play modules for LASNet, namely CLM, CAM, and ESM, which are responsible for object location, region activation and edge sharpening, respectively.
In these three modules, we focus on extracting the complementary information between the two feature combinations of multiplication and summation rather than directly extracting the complementary information between RGB and TIR features, which is different from previous methods.
The tight integration of the paradigm and the three modules results in our complete solution.
Comprehensive ablation studies demonstrate the effectiveness of the above three modules, and extensive comparisons demonstrate the superiority of our LASNet.



\ifCLASSOPTIONcaptionsoff
  \newpage
\fi

\bibliographystyle{IEEEtran}
\bibliography{RGBTSSref}

%



%

\end{document}